\documentclass{article}

\pdfoutput=1 

\usepackage{arxiv}

\usepackage[utf8]{inputenc} 
\usepackage[T1]{fontenc}    
\usepackage[colorlinks=true, citecolor=cyan]{hyperref}       
\usepackage{url}            
\usepackage{booktabs}       
\usepackage{amsfonts}       
\usepackage{nicefrac}       
\usepackage{microtype}      
\usepackage{lipsum}		
\usepackage{graphicx}
\usepackage{caption}
\usepackage{textcomp}

\usepackage{xcolor}
\usepackage[normalem]{ulem}
\hypersetup{colorlinks,urlcolor=blue}

\makeatletter
\DeclareUrlCommand\ULurl@@{%
  \def\UrlLeft{\uline\bgroup}%
  \def\UrlRight{\egroup}}
\def\ULurl@#1{\hyper@linkurl{\ULurl@@{#1}}{#1}}
\DeclareRobustCommand*\ULurl{\hyper@normalise\ULurl@}
\makeatother

\title{Utilizing a null class to restrict decision spaces and defend against neural network adversarial attacks}

\author{ Matthew J.~Roos\thanks{\href{https://binarycognition.com/mattroos/}{https://binarycognition.com/mattroos/}} \\
	Binary Cognition, LLC\\
	Denver, CO\\
	\texttt{matt.roos@binarycognition.com} \\
}

\renewcommand{\headeright}{}
\renewcommand{\undertitle}{Technical Report}

\begin{document}
\maketitle

\begin{abstract}
Despite recent progress, deep neural networks generally continue to be vulnerable to so-called \emph{adversarial examples}---input images with small perturbations that can result in changes in the output classifications, despite no such change in the semantic meaning to human viewers. This is true even for seemingly simple challenges such as the MNIST digit classification task. In part, this suggests that these networks are not relying on the same set of object features as humans use to make these classifications. In this paper we examine an additional, and largely unexplored, cause behind this phenomenon---namely, the use of the conventional training paradigm in which the entire input space is parcellated among the training classes. Owing to this paradigm, learned decision spaces for individual classes span excessively large regions of the input space and include images that have no semantic similarity to images in the training set. In this study, we train models that include a \emph{null} class. That is, models may “opt-out” of classifying an input image as one of the \emph{digit} classes. During training, null images are created through a variety of methods, in an attempt to create tighter and more semantically meaningful decision spaces for the digit classes. The best performing models classify nearly all adversarial examples as nulls, rather than mistaking them as a member of an incorrect digit class, while simultaneously maintaining high accuracy on the unperturbed test set. The use of a null class and the training paradigm presented herein may provide an effective defense against adversarial attacks for some applications. Code for replicating this study will be made available at \href{https://github.com/mattroos/null_class_adversarial_defense}{https://github.com/mattroos/null\_class\_adversarial\_defense}.
\end{abstract}


\section{Introduction}

Shortly after the resurgence of deep neural networks as machine learning models for object and image classification \cite{krizhevsky2012imagenet, lecun2015deep} it was realized that modest image perturbations could be made such that these models incorrectly classify an image, with high classification probability \cite{szegedy2013intriguing}. In some case these perturbations are imperceptible to humans, suggesting that these models do not rely on the same causal features as humans for visual perception, and that decision boundaries between classes are extremely close to some image samples. In this work we address the latter issue, speculating as to the reason for its existence, and investigating a training approach that attempts to mitigate the problem.

Imagine training a child to classify images of squares and diamonds as belonging to their respective classes. If we then show the child a picture of a circle, an eight-pointed star, or white noise, they are likely to declare the object to be neither a square or a diamond---a perfectly acceptable response. Yet when it comes to machine learning models, this is precisely the behavior we do not allow the conventional model to exhibit under such ambiguity. That is, the model is trained on N classes and the model architecture allows N outputs. The model is forced to parcellate the input space into N decision spaces that together fill the entire input space. However, for most models and tasks, true samples (whether used for training or not) exist in only a small subset of the input space. We speculate that this situation, coupled with the high dimensionality of the input space (e.g., $224\times224=50176$ dimensions for the typical ImageNet model), results in samples near decision boundaries that are orthogonal to latent dimensions with no semantic meaning. In the following section we present results from a simple toy problem designed to allow us to visualize this dilemma.

Although the challenge of defending against adversarial examples is most often conveyed with photographic imagery like that of ImageNet, it persists in simpler tasks and smaller models as well, such as MNIST. While progress has been made, even what is arguably the most successful defense by a \emph{conventional CNN} model \cite{madry2017towards} has been shown to be susceptible to certain types of adversarial images \cite{schott2018towards}. There has been greater success when using other model forms, however. Using an \emph{analysis by synthesis} approach, Schott and colleagues \cite{schott2018towards} demonstrate an MNIST model that is highly effective in defending against adversarial examples, and produces digit predictions that align well with human perception---i.e., classification "mistakes" on adversarial examples occur when the perturbations push the image toward one of a different class, as perceived by humans, such as altering the upper portion of a 4 so that it appears more similar to a 9. Yet, we note that their model is unlike a conventional feedforward CNN---rather, it uses gradient descent during inference to find an optimal solution based on maximizing the lower bound on the log-likelihood for each class. While the approach provides good results, it is computationally very expensive in comparison with conventional CNN models, and is unlikely to provide a practical defense against adversarial attacks on more challenging tasks, under currently available hardware.

The models we investigate in this report have architectures that allow N+1 outputs, for a task with N object classes. This N+1\textsuperscript{th} class is a \emph{null} class that is conceptually equivalent to a "no-decision" by the model. We find that these models, when trained with null samples in addition to the standard training samples, classify unperturbed images nearly as well as their baseline counterparts yet are adept at detecting perturbed images---thereby provided a computationally affordable way to defend against adversarial attacks.

\section{A motivational example}
\label{sec:headings}

In this section we present a demonstrative toy model and two methods of training---a conventional method and one that includes null samples. We highlight the characteristics of the learned decision spaces within the input space, and the susceptibility of the conventionally-trained model to adversarial attacks.

\begin{figure}
	\centering
	\includegraphics[scale=0.5]{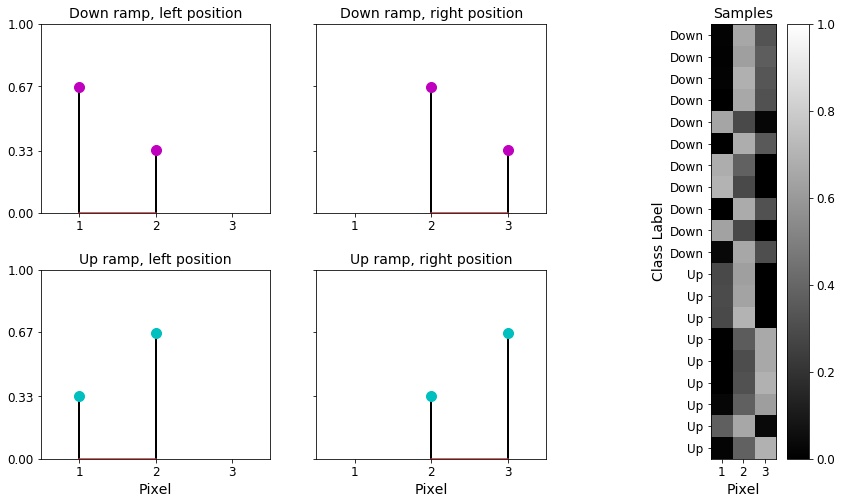}
	\caption{\emph{Left}: Prototype “images” of two classes of objects (down ramps and up ramps) that may be embedded in the three-pixel input space at one of two locations, for a total of four prototype images. \emph{Right}: A random selection of training images (rows). Training images are created by adding uniform noise to prototype images, and then clipping the values to maintain a range of [0, 1] for all pixels.}
	\label{fig:fig_ramp_samples.png}
\end{figure}

Our toy environment is a linear array of three pixels and our toy classes are composed of two adjacent pixels for two classes—a down ramp and an up ramp. Examples of objects with these classes are shown in Figure~\ref{fig:fig_ramp_samples.png}. The sizes of the objects and input space were chosen such that (1) the toy model can be a convolutional neural network (CNN), the model type utilized in most research studies on adversarial attacks, and (2) the decision spaces within the 3D input space can easily be visualized.

\begin{figure}
	\centering
	\includegraphics[scale=0.4]{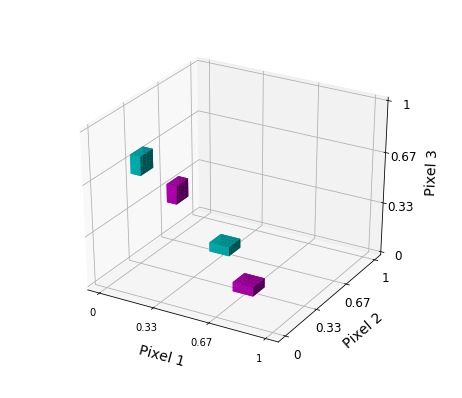}
	\caption{The optimal decision spaces within the input space for the down ramps (magenta) and the up ramps (cyan). In other words, the spaces that completely contain all ramp-containing input images, as defined in the text, but no other input images.}
	\label{fig:fig_decision_space_optimal}
\end{figure}

As shown in Figure~\ref{fig:fig_ramp_samples.png}, the prototype objects have pixel values of (2/3, 1/3) and (1/3, 2/3) for the down and up ramps, respectively, and a ramp may exist in the left two pixels or the right two pixels of the three-pixel image. During training, we create training samples by adding uniform noise of range [-0.05, 0.05] to the prototype images in each dimension, then clipping the results to maintain input values in [0, 1]. Thus, the optimal decision spaces of a model are those depicted in Figure~\ref{fig:fig_decision_space_optimal}---optimal in the sense that the spaces encapsulate the entire space of possible input images which contain one of the two classes of objects, as defined above, and do not encapsulate input images outside of this space.

We constructed a simple CNN model for classifying down and up ramps. The layer-wise ordering is identical that of our MNIST models (Table~\ref{tab:table_architecture}). The model was comprised of, in order: two consecutive convolutional layers with 2-pixel kernels (8 output channels and 16 output channels), a max pooling layer, a dense layer with 32 outputs, and a dense layer with 3 outputs—two for the two ramp classes, and one for the null class. The architecture is unlikely to be pertinent to the forthcoming results, but was chosen simply because it is the same as that of the MNIST CNN architecture we used for other experiments, albeit with fewer channels and nodes.

We first trained the model in the conventional manner, using only training images that contained one of the two ramp objects (no null samples). We trained on 20 epochs of 1000 samples, with a batch size of 32 images. Convergence happened quickly, with near-perfect accuracy achieved after three epochs. We then probed the trained model, creating input images that grid-sampled the 3D space, and determining the models output prediction for each of those inputs.

\begin{figure}
	\centering
	\includegraphics[scale=0.43]{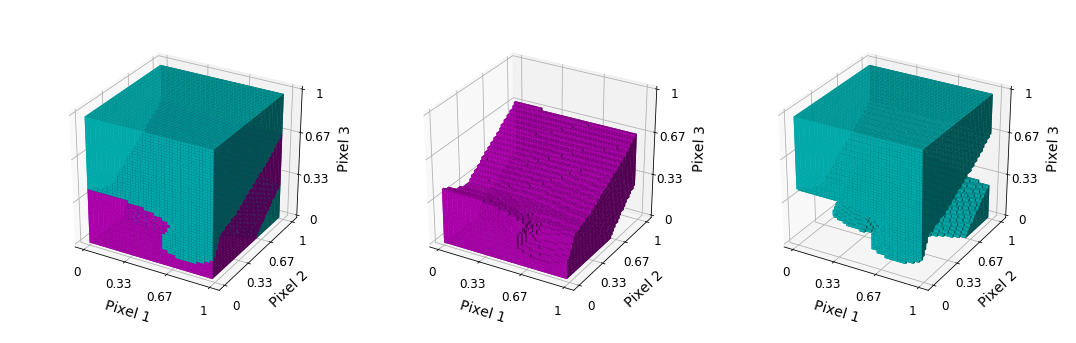}
	\caption{Decision spaces for a CNN model applied to the toy task, when trained in the conventional manner, without null samples. Together the two decision spaces cover the entire input space, despite there being only small subspaces that contain true up and down ramp images (see Figure \ref{fig:fig_decision_space_optimal}). \emph{Left}: The down (magenta) and up (cyan) decision spaces. \emph{Middle}: The down decision space alone. \emph{Right}: The up decision space alone.}
	\label{fig:fig_decision_space_baseline}
\end{figure}

Figure~\ref{fig:fig_decision_space_baseline} shows the results. As must be the case under this type of training, the decision spaces for the two classes collectively span the entire input space, despite the fact that only a small subspace of each space contains representative samples of its respective class. Importantly, the decision borders abut one another, and many locations of the up space lie close to the locations of true down samples (compare Figures \ref{fig:fig_decision_space_optimal}~and~\ref{fig:fig_decision_space_baseline}), and vice versa. In other words, the model is clearly amenable to adversarial attacks.

\begin{figure}
	\centering
	\includegraphics[scale=0.43]{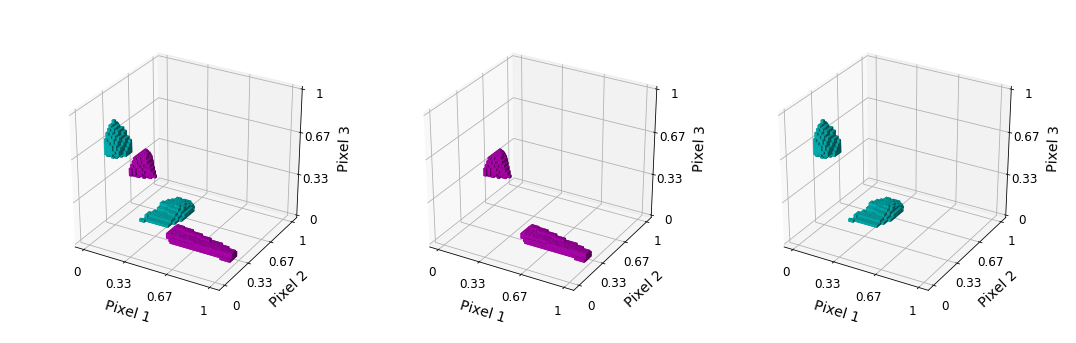}
	\caption{Decision spaces for a CNN model applied to the toy task, when trained with both ramp samples and with uniform-noise null samples. Only the decision spaces for the down and up ramp classes are shown. All the remaining regions of the input space constitute the null decision space. Decision spaces for the two ramp classes tightly constrain the subspaces of true up and down ramp images, in contrast to when null samples are not used in training (Figure~\ref{fig:fig_decision_space_baseline}). \emph{Left}: the down (magenta) and up (cyan) decision spaces. \emph{Middle}: The down decision space alone. \emph{Right}: the up decision space alone.}
	\label{fig:fig_decision_space_uniform}
\end{figure}

We then trained the model using uniform-noise null samples in addition to ramp samples. During training, 50\% of the up and down ramp samples were replaced with noise samples, with pixel values randomly chosen in [0, 1]. The null samples were labeled as such (not as up or down samples). In contrast to the fast convergence under conventional training, the model was trained for 30 epochs and was still converging at the end of training. However, very high accuracy on up and down ramp samples was achieved early in the training, suggesting that much of the training period involved tightening of the up and down decision spaces such that null samples would not be misclassified as ramps. Figure~\ref{fig:fig_decision_space_uniform} shows the ramp decision spaces of the model at the end of this training period. In contrast to those of conventional training, they tightly encompass the true up and down ramp samples, leaving the remaining space as the null class decision space. Thus, creating perturbed images that will be misclassified by this model is nearly impossible, except when the perturbed image truly takes on the characteristics of the class predicted by the model—i.e., no longer looking like an object from the class of the source image used to create the perturbed image.

Having demonstrated how conventional training might lead to decision spaces that are susceptible to adversarial attacks, we hypothesize that this same scenario may hold true for MNIST (and more complex) tasks and models. We explore this in the remainder of the paper.

\section{Network architecture}
\label{sec:architecture}

All models trained and tested in this study have the same architecture—a typical CNN of size comparable to most modern networks trained for the MNIST classification task. The specific architecture is presented in Table~\ref{tab:table_architecture}. Given that our goal was to explore methods of defending against adversarial attacks, we chose this conventional, well-performing architecture rather than a particular “best” architecture, as measured by accuracy on the test set, or via other metric. Our baseline model scored and average accuracy of 99.23\% accuracy on the MNIST test set.

The only notable difference between our model’s architecture and that of typical MNIST CNNs is the use of eleven output nodes rather than ten, to accommodate the null class in addition to the digit classes, 0 through 9.

\begin{table}
	\caption{MNIST network architecture}
	\centering
	\begin{tabular}{lllll}
		\toprule
		Layer type     & Input size     & Kernel size   & Rate & Output size \\
		\midrule
		Conv2D & (28, 28, 1)    & (3, 3, 1)  &  & (26, 26, 32) \\
		ReLU     & (26, 26, 32)  &  &  &  \\
		Conv2D & (26, 26, 32)  & (3, 3, 1)  &  & (24, 24, 64) \\
		ReLU     & (24, 24, 64)  &   &  &  \\
		MaxPool2D & (24, 24, 64) & (2, 2) & & (12, 12, 64) \\
		Dropout & (12, 12, 64) & & 0.25 & (12, 12, 64) \\
		Flatten & (12, 12, 64) & & & 9216 \\
		Dense & 9216 & & & 128 \\
		Dropout & 128 & & 0.5 & 128 \\
		Dense & 128 & & & 11 \\		
		\bottomrule
	\end{tabular}
	\caption*{The architecture of all the MNIST-trained models, from input (top) to output (bottom). Layer type names (abbreviations) are those of the Keras API for TensorFlow 2.1. Dropout layers were applied only during training. The layer structure of the toy model was identical, but with different input, output, and kernel sizes (see text).}
	\label{tab:table_architecture}
\end{table}

\section{Training methods}
\label{sec:training_methods}

\subsection{All models}

Training (and testing) sample images were input to the models as a 28\texttimes28 matrices of 32-bit floating point numbers, with a range of [0, 1]. Models were trained for 20 epochs with a batch size of 32 images. Here the term \emph{epoch} refers to 60,000 training samples, but this does not map one-to-one to the 60,000 images in the MNIST training set. Instead, training samples were drawn randomly and independently from the training set for each batch, and thus a single epoch could exclude some training set images entirely, and could include repeats of unique training images. MNIST training images \emph{were not} augmented when used as digit training samples. MNIST training images \emph{were} altered when used to create some types of null samples (see below), but not in any manner similar to that traditionally done for training set augmentation (e.g., translation or noise-addition).

For training that included 50\% null samples (see below), the number of null and digit (non-null) samples in individual batches was randomly determined by sampling from a uniform distribution such that the expected number of null and digit samples was each equal to half of the batch size. Thus, in a given epoch the total expected number of null and digit samples was 30,000 each. When multiple types of null samples were used for training, the type of sample for a given null sample was randomly chosen with equal probability across types.

Neuron biases were initialized with zeros and weights were randomly initialized using the Glorot uniform method \cite{glorot10understandingthe}. We used trained the models using cross-entropy loss and the Adam gradient descent optimizer \cite{kingma2014adam} with the following parameter values: $\alpha=0.001$, $\beta_1=0.9$, $\beta_2=0.999$, $\epsilon=10^{-08}$.

Many of the above details of the training regimen where chosen for coding convenience or reduction of training durations, not for model performance reasons. For example, although not systematically tested, we suspect that precise choices for batch size, parameter initialization, or optimization meta-parameters will not significantly impact any of the findings reported herein.

\subsection{Baseline models}

MNIST models typically have ten output nodes, one for each of digit classes. Our \emph{baseline} models include an 11\textsuperscript{th} output node that corresponds to the null class. However, no null samples were used in training of baseline models and thus it is very unlikely that the models will generate a null decision, regardless of the input. During testing on the MNIST test set, as well as on null samples, we never observed a null decision by a baseline model.

\subsection{Null-trained models}

\emph{Null-trained} models are those that were trained with null samples in addition to images from the MNIST training set. On average, 50\% of the training samples where null samples and 50\% of the training samples were digit samples (and thus roughly 5\% of training samples belonged to an individual digit class). A much larger percentage of null samples than individual digit samples was chosen because of the vastly larger ground truth space of the null class (see below). We used three types of null samples, and trained models using all possible combinations of these types (resulting in seven different null-trained models). In general, the goal of training with null samples is to constrain the decision spaces of digit classes such that they generalize well to test samples, while classifying all other samples in the input space as null samples, thereby mitigating the impact of adversarial examples. However, the input space is vast, even for small images such as those of MNIST. The possible number of inputs for 28\texttimes28 images of 8-bit integers is:

\medskip
\centerline{$2^{8^{28\times28}} = 256^{784} = {10^{2.40824}}^{784} > 10^{1888}$}
\smallskip

Thus, efficient sampling of the null samples from this space is imperative for success. Below, we describe the three types of null samples used in training, and provide our reasoning for their selection. Examples of the null types are shown in Figure~\ref{fig:fig_null_image_examples}.

\subsubsection{Uniform-noise null samples}

Uniform noise samples are images composed of \emph{i.i.d.}~pixel samples taken from a uniform distribution. They were chosen because they provide broad, albeit sparse, sampling of the input space, most of which is null space. There is a small overlap with the input space of ground truth digit decision spaces, but that overlap is exceptionally small, and randomly created sample are extremely unlikely to fall within digit spaces.

\subsubsection{Mixed-digit null samples}

As suggested by the demonstrative toy model (Figure~\ref{fig:fig_decision_space_baseline}), there may exist regions of ground truth decision spaces that are in close proximity to each other in Euclidean space, despite being a great distance apart in some latent semantic space. Because conventional models fill the entire input space with digit decision spaces, it seems likely that adversarial images may cross digit class boundaries at these Euclidean-proximal locations, and thus be misclassified. To promote the learning of null spaces that create a Euclidean separation between digit decision spaces, we create null samples that are simply the mean of two samples of different digit classes.

\subsubsection{Shuffled-digit null samples}

Finally, we create null samples from individual digit samples by tiling them at different scales and randomly shuffling the tiles to create a new image. Tile sizes for image are randomly chosen from {1\texttimes1, 2\texttimes2, 4\texttimes4, 7\texttimes7, and 14\texttimes14 . All images within a batch are tiled at the same scale and shuffled in the same order. The goal of these null samples is to maintain some of the higher-order statistical properties of the training images, thereby creating samples that may be closer to the digit samples along dimensions that most uniform or mixed null samples do not lie. For larger tile sizes in particular, the shuffled null samples have large-scale components that are very similar to those of digit class images yet lack the semantic meaning of digit classes. This type of null sample is also inspired by the observation that many trained CNNs make decisions based on statistical properties of small regions within an image (e.g., texture of fur or skin) rather than holistic or composite properties of the object within the image \cite{ballester2016ontheperformance, brendel2018approximating, geirhos2018imagenet}. Using shuffled images as null samples may force the model to learn to make decisions based on large-scale object composition more so than the collection of local features.

\begin{figure}
	\centering
	\includegraphics[scale=0.5]{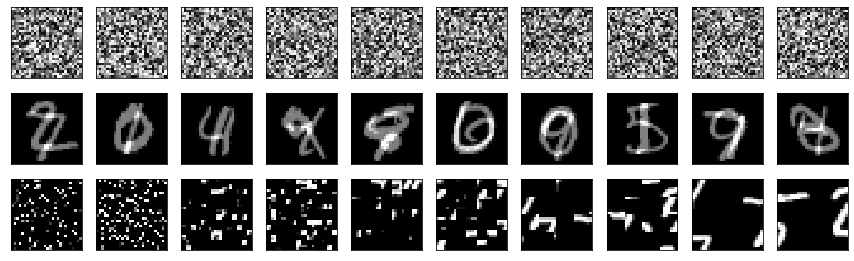}
	\caption{Examples of null image samples. \emph{Top}: Uniform noise samples. \emph{Middle}: Mixed-digit samples. \emph{Bottom}: Shuffled-digit samples, with tiles sizes of 1, 2, 4, 7, and 14, from left to right, with two images per tile size.}
	\label{fig:fig_null_image_examples}
\end{figure}

\section{Results}

Using the training methods described above, we trained 30 models for each model type, which included a baseline model type and a null-trained model type for all possible combinations of null image types. Thus, we obtained 30 instantiations of eight different models. For each “set” of eight models of a given training run, the random number generator seed was set to a different value (1 of 30 seeds).

For each of 30 sets of eight models we: (1) measured the accuracy of the models on the MNIST test set, (2) used the baseline model to create 1000 adversarial images from the MNIST test set, (2) scored the null-trained models (hereafter referred to simply as the null models) on the adversarial images, and (3) scored the baseline and null models on perturbed images generated over a large range of epsilon values (epsilon being the scalar multiplier of the noise (gradient) image added to the source image when creating adversarial images---not to be confused with the epsilon term of the Adam gradient descent optimizer).

\subsection{MNIST test set scores for all models}

Before testing models on adversarial images, we measured accuracy on the unperturbed MNIST test set images. As seen in Figure~\ref{fig:fig_test_set_scores}, accuracy is approximately 99\% for all model types, with the null models performing slightly worse than the baseline models. This modest accuracy drop is likely due to the tighter decision spaces of the null models versus the baseline models---lowering null model generalizability but also lowering the likelihood of misclassification of adversarial images, as will be shown.

\begin{figure}
	\centering
	\includegraphics[scale=0.6]{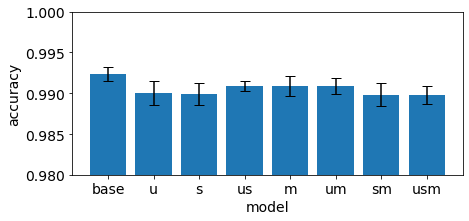}
	\caption{Model accuracy scores on unperturbed MNIST test images. Bar heights and error bars are means and standard deviations, respectively, computed across all 30 sets of trained and tested models. The baseline model is labeled \emph{base} while the null models are labeled with combinations of \emph{u}, \emph{s}, and \emph{m}, indicating the types of null images on which the models were trained—uniform, shuffled, and mixed, respectively.}
	\label{fig:fig_test_set_scores}
\end{figure}

\subsection{Adversarial examples at epsilon threshold}

For this study we used the fast gradient sign method \cite{goodfellow2014explaining} (FGSM) for generating adversarial images. When using FGSM, the signed gradient image is multiplied by a scalar value, added to the source image from which the gradient was computed, and the composite is clipped to maintain the range of values of the original MNIST digits (typically scaled to [0, 1], as in this study). Several research groups \cite{madry2017towards, zheng2018distributionally, schott2018towards} have explored adversarial images created at epsilon=0.3, presumably finding that for this value (1) many of the perturbed images will be misclassified by the source (baseline) model, and (2) a large subset of those perturbed images continue to be perceived by humans as digits of the same class as the source images from which they were derived (albeit with perceptible noise in many cases).

Rather than selecting a single epsilon value when generating adversarial images, we instead computed the signed gradient image and then searched for the minimum epsilon value (within tolerance of 10\textsuperscript{-3}) that resulted in a misclassification by the source model for that specific gradient and source image. Thus, we found the epsilon threshold for each image, and all adversarial images created in this manner lie very close to a boundary between decision spaces.

In Figure~\ref{fig:fig_adversarial_samples_baseline}, we display 30 examples of the 1000 adversarial images created for use with a single set of eight models (baseline plus null models). A different set of 1000 adversarial examples is created for each of the 30 sets of eight models. Most of the perturbed images are perceptually identifiable as belonging to the digit class of the source image, with varying degrees of detectable noise. Because the images were created just above epsilon threshold, the probability scores of the predicted class given by the model are relatively low (often near 0.5). To see the impact of slightly larger epsilon values on both the image quality and resulting probability scores, we created a set of perturbed images identical to those of Figure~\ref{fig:fig_adversarial_samples_baseline} except epsilon values were increased by a factor of 1.5, thereby moving them further away from decision boundaries in most cases. As seen in Figure~\ref{fig:fig_adversarial_samples_baseline_1p5}, compared to those of Figure~\ref{fig:fig_adversarial_samples_baseline}, image perturbations are more apparent although most images are still clearly identifiable (by humans) as the digit class from which they were derived. Probability scores generated by the source model are much higher, however---in line with those of the most poignant adversarial examples highlighted by the machine learning community.

\begin{figure}
	\centering
	\includegraphics[scale=0.5]{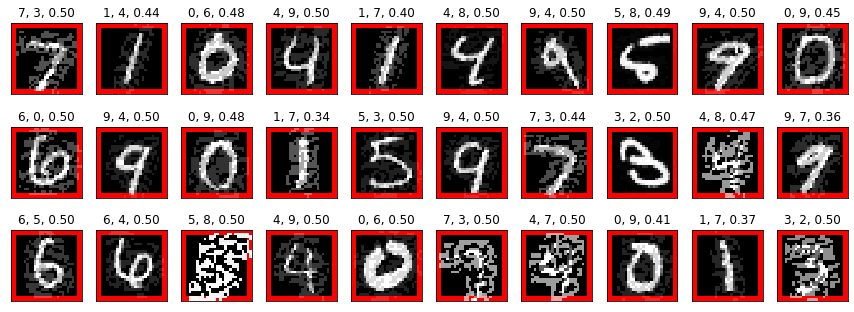}
	\caption{Examples of adversarial images created from a baseline model. Each image was created with the smallest epsilon value (epsilon threshold) possible such that the source model misclassified the image. Images that are misclassified are given a red border (which is all of them in this figure, by design). Numbers above each plot are, in order: source image label, model prediction label, and score (probability) for the predicted label.}
	\label{fig:fig_adversarial_samples_baseline}
\end{figure}

\begin{figure}
	\centering
	\includegraphics[scale=0.5]{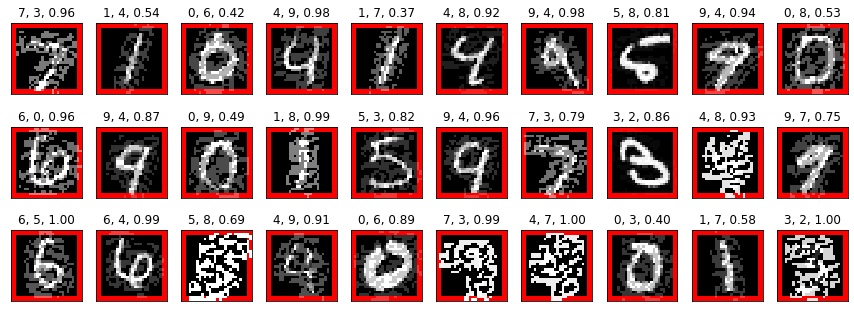}
	\caption{Perturbed images identical to those in Figure 2, except with epsilon values 1.5 times larger. Images appear slightly noisier although often remain clearly identifiable to humans. Probability scores of predicted classes are much larger, however, often reaching 0.9 or higher. The source model misclassified all the images, as indicated by the red borders. Plots are titled in the same manner as in Figure~\ref{fig:fig_adversarial_samples_baseline}.}
	\label{fig:fig_adversarial_samples_baseline_1p5}
\end{figure}

Despite intentionally creating adversarial images with the minimum perturbation possible, many still have an epsilon value near 0.3. As seen in Figure~\ref{fig:fig_epsilon_thresholds_hist}, the distribution of epsilon values across all epsilon-threshold adversarial images (30,000 in total) peaks near 0.2, with more than 80\% of the values falling below 0.4.

\begin{figure}
	\centering
	\includegraphics[scale=0.55]{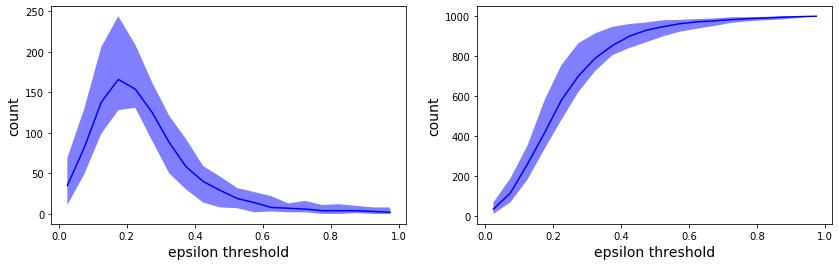}
	\caption{Distribution of epsilon values at threshold. A composite histogram (left) and composite cumulative histogram (right) are plotted for the adversarial images created from all 30 baseline models. Individual histograms and cumulative histograms were created for each baseline model’s adversarial images, with a bin size of 0.05. Displayed curves are the median counts across the 30 sets of adversarial images, with shaded areas indicating the minimum and maximum values.}
	\label{fig:fig_epsilon_thresholds_hist}
\end{figure}

\setcounter{footnote}{0} 

Having created adversarial images for each baseline model, we then tested those images on the null models of the same set (seven null models)\footnote{There is nothing that naturally groups our trained models into sets of eight (one baseline and seven null models), except that the eight models were trained sequentially inside a for loop, which was initialized with a unique random number generator seed. Thus, adversarial images could have been tested on all 7\texttimes30 = 210 null models, rather than testing adversarial images only on null models within the same set as the baseline model from which the image were created. This was not done, however, for the sake of computational expediency.}, determining the percentage of adversarial images that were classified correctly, misclassified (as a digit class other than that of the source image), or unclassified (predicted to be in the null class). Results are shown in Figure~\ref{fig:fig_scores_at_threshold}. As desired, misclassifications drop to near zero when models are trained on a variety of null images in addition to digit images. When training with certain types of null images, most of the epsilon-threshold adversarial images are instead classified as null images, and a small number are correctly classified, in contrast to results from the baseline model.

\begin{figure}
	\centering
	\includegraphics[scale=0.6]{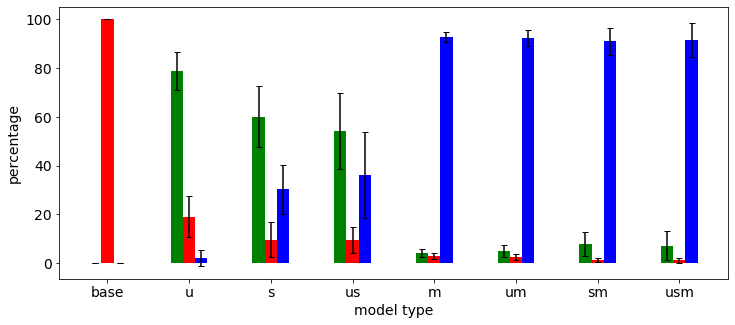}
	\caption{Model scores on adversarial images at epsilon threshold. \emph{Green}: classified correctly. \emph{Red}: misclassified (classifying an image as a digit class other than that of the source image). \emph{Blue}: unclassified (a null classification). Bar height are means and error bars are standard deviations, computed across the 30 sets of eight models. The baseline model is labeled \emph{base} while the null models are labeled with combinations of \emph{u}, \emph{s}, and \emph{m}, indicating the types of null images on which the models were trained---uniform, shuffled, and mixed, respectively.}
	\label{fig:fig_scores_at_threshold}
\end{figure}

Examples of null model predictions and scores for specific images are shown in Figure~\ref{fig:fig_adversarial_samples_u} through Figure ~\ref{fig:fig_adversarial_samples_usm}. All images are the same as those of Figure~\ref{fig:fig_adversarial_samples_baseline}. As the variety of null image types on which the null model was trained increases, models increasingly classify adversarial examples as null/unclassified, rather than misclassifying them---a trend also seen in the collective results reported in Figure~\ref{fig:fig_scores_at_threshold}.

\begin{figure}
	\centering
	\includegraphics[scale=0.5]{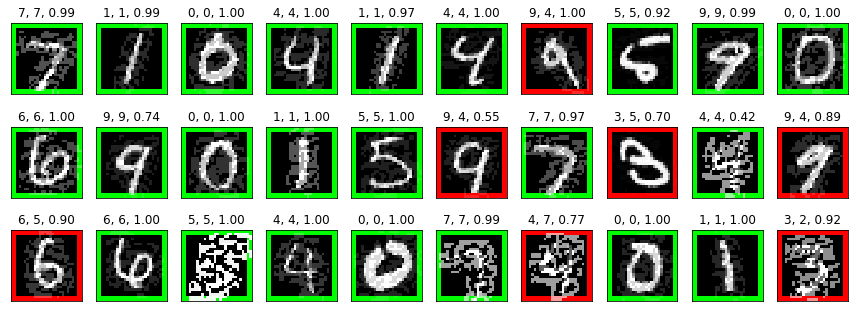}
	\caption{Example adversarial images from Figure~\ref{fig:fig_adversarial_samples_baseline}, scored by a model trained with \emph{uniform} noise null images. Colors and plot titles are as those in Figures~\ref{fig:fig_scores_at_threshold}~and~\ref{fig:fig_adversarial_samples_baseline}, respectively.}
	\label{fig:fig_adversarial_samples_u}
\end{figure}

\begin{figure}
	\centering
	\includegraphics[scale=0.5]{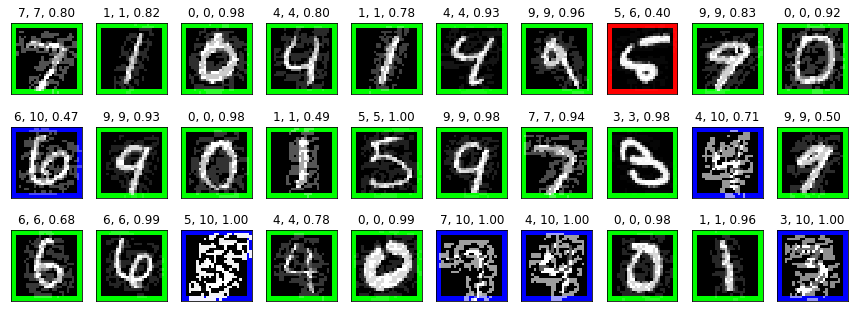}
	\caption{Example adversarial images from Figure~\ref{fig:fig_adversarial_samples_baseline}, scored by a model trained with \emph{shuffled} noise null images. Colors and plot titles are as those in Figures~\ref{fig:fig_scores_at_threshold}~and~\ref{fig:fig_adversarial_samples_baseline}, respectively.}
	\label{fig:fig_adversarial_samples_s}
\end{figure}

\begin{figure}
	\centering
	\includegraphics[scale=0.5]{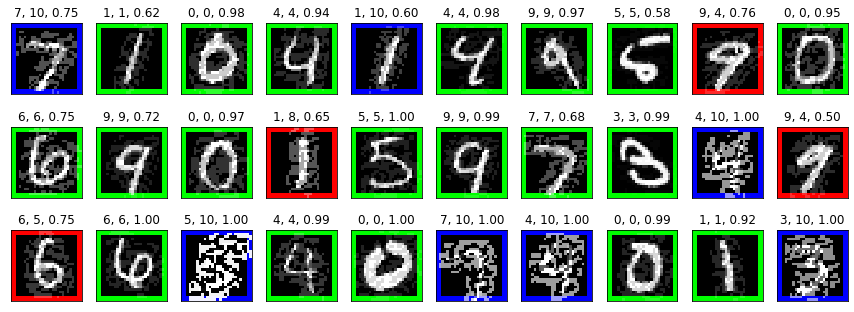}
	\caption{Example adversarial images from Figure~\ref{fig:fig_adversarial_samples_baseline}, scored by a model trained with \emph{uniform} noise null images and \emph{shuffled} null images. Colors and plot titles are as those in Figures~\ref{fig:fig_scores_at_threshold}~and~\ref{fig:fig_adversarial_samples_baseline}, respectively.}
	\label{fig:fig_adversarial_samples_us}
\end{figure}

\begin{figure}
	\centering
	\includegraphics[scale=0.5]{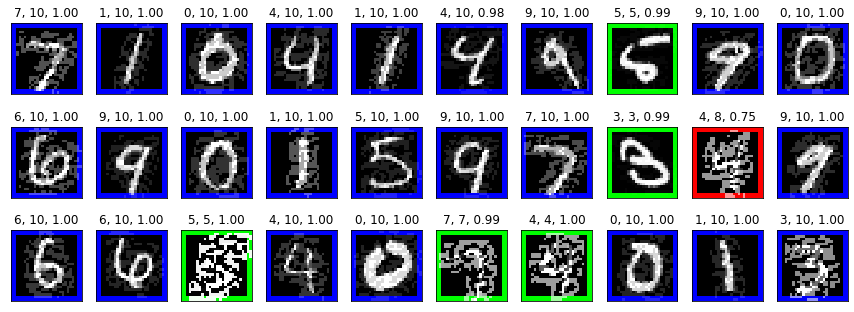}
	\caption{Example adversarial images from Figure~\ref{fig:fig_adversarial_samples_baseline}, scored by a model trained with \emph{mixed-class} null images. Colors and plot titles are as those in Figures~\ref{fig:fig_scores_at_threshold}~and~\ref{fig:fig_adversarial_samples_baseline}, respectively.}
	\label{fig:fig_adversarial_samples_m}
\end{figure}

\begin{figure}
	\centering
	\includegraphics[scale=0.5]{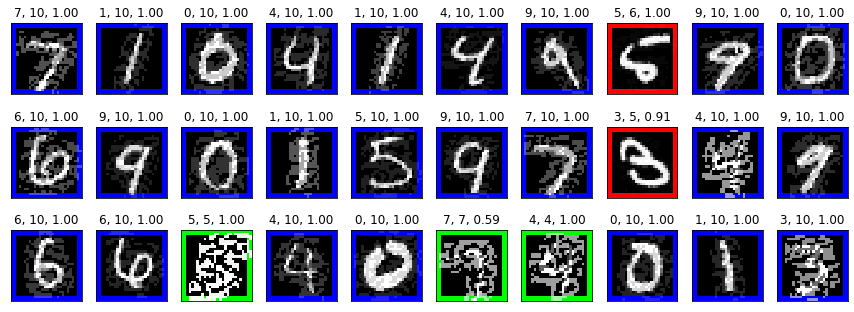}
	\caption{Example adversarial images from Figure~\ref{fig:fig_adversarial_samples_baseline}, scored by a model trained with \emph{uniform} noise null images and \emph{mixed-class} null images. Colors and plot titles are as those in Figures~\ref{fig:fig_scores_at_threshold}~and~\ref{fig:fig_adversarial_samples_baseline}, respectively.}
	\label{fig:fig_adversarial_samples_um}
\end{figure}

\begin{figure}
	\centering
	\includegraphics[scale=0.5]{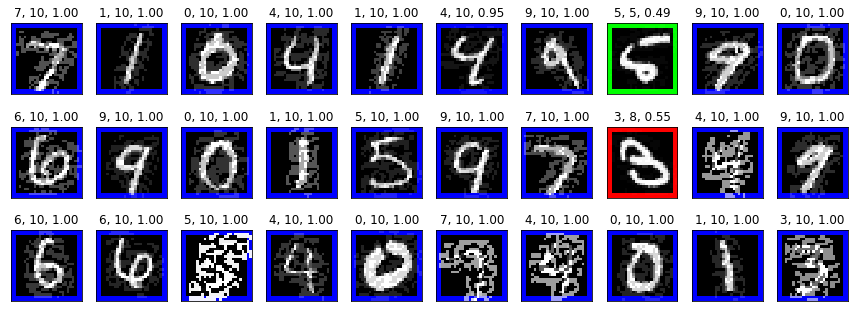}
	\caption{Example adversarial images from Figure~\ref{fig:fig_adversarial_samples_baseline}, scored by a model trained with \emph{shuffled} null images and \emph{mixed-class} null images. Colors and plot titles are as those in Figures~\ref{fig:fig_scores_at_threshold}~and~\ref{fig:fig_adversarial_samples_baseline}, respectively.}
	\label{fig:fig_adversarial_samples_sm}
\end{figure}

\begin{figure}
	\centering
	\includegraphics[scale=0.5]{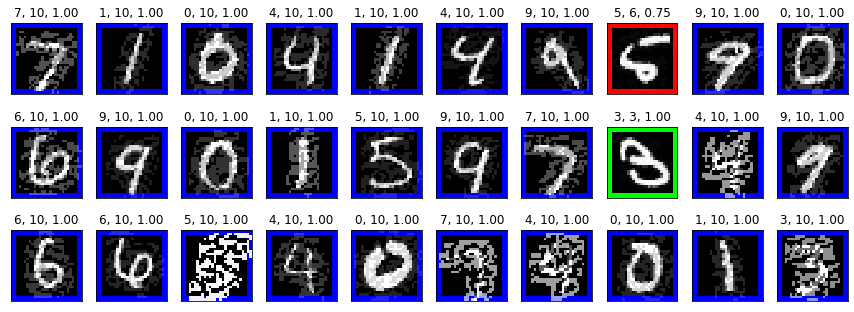}
	\caption{Example adversarial images from Figure~\ref{fig:fig_adversarial_samples_baseline}, scored by a model trained with \emph{uniform} noise null images, \emph{shuffled} null images, and \emph{mixed-class} null images. Colors and plot titles are as those in Figures~\ref{fig:fig_scores_at_threshold}~and~\ref{fig:fig_adversarial_samples_baseline}, respectively.}
	\label{fig:fig_adversarial_samples_usm}
\end{figure}

\subsection{Adversarial examples across a range of epsilon values}

We next examined model performances across a larger range of perturbations, systematically varying epsilon from 0 to 1, generating a set of perturbed images for each epsilon value, and scoring the models on those images. Similar to the previous epsilon-threshold testing, 1000 perturbed images were created from each baseline model, and tested on all models within the same set. Because the epsilon values were varied, the baseline model itself was tested on these perturbed images, as well as the null models.

Results are summarized in Figure~\ref{fig:fig_scores_versus_epsilon}. As expected, the baseline models score perfectly when epsilon is equal to zero (only MNIST test set images that were given a correct classification by a baseline model were considered as source images for the creation of perturbed images), but scores fall to nearly zero when epsilon is equal to one. It is worth noting that although an epsilon value of one is greater than the epsilon threshold for all the perturbed images, this does not mean that the baseline model misclassification error rate must reach 100\% at high epsilon. This is because for some perturbed images, misclassification occurs at epsilon threshold (by definition), but reverts to the correct classification as epsilon is increased further.

We briefly note three general null model trends exhibited in Figure~\ref{fig:fig_scores_versus_epsilon}, and reserve further comments for the Discussion section. First, the null models trained solely on uniform-noise null samples show relatively weak resistance to adversarial images, performing only slightly better than the baseline models. Second, null models trained with null samples that included mixed-class null images robustly defend against perturbed images for epsilon values up to roughly 0.6---a range that covers nearly all of the epsilon-threshold adversarial examples. Here we use the word defend to mean that images are either classified correctly or given a null classification, rather than being misclassified as a digit other than that of the source image. The defense begins to breakdown as epsilon increases above 0.6. Finally, null models trained with null samples that included shuffled images robustly defend against perturbed images with epsilon values roughly 0.5 and above. When trained on both mixed-class images and shuffled images, null models robustly defend against perturbed images across the full range of tested epsilon values.

\begin{figure}
	\centering
	\includegraphics[scale=0.5]{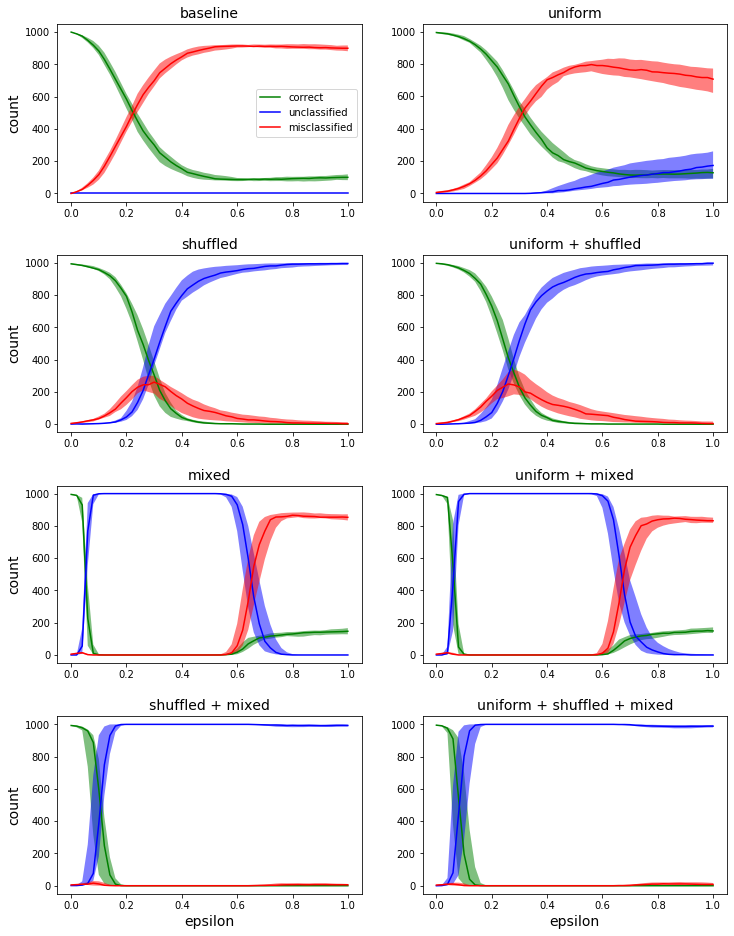}
	\caption{Null models defend against adversarial images across a range of epsilon values. Each panel summarizes the predictions made by the specified model type (panel titles indicate the type of null images used during training) as a function of epsilon, for a set of perturbed images created from the baseline model within each set of models. \emph{Green}: classified correctly. \emph{Red}: misclassified. \emph{Blue}: unclassified (a null classification). Displayed curves are the median counts across the 30 sets of perturbed images, with shaded areas indicating the 25\textsuperscript{th} and 75\textsuperscript{th} percentiles.}
	\label{fig:fig_scores_versus_epsilon}
\end{figure}

To highlight the performance difference between the baseline models and the null models trained with shuffled and/or mixed-class images, we replot some of the Figure~\ref{fig:fig_scores_versus_epsilon} data in Figure~\ref{fig:fig_scores_versus_epsilon_model_compare}. Therein, one can directly compare model misclassification scores. As seen in the rightmost panel, the best performing null models make almost no misclassification errors. However, they do not generalize as well as the baseline models—giving a null classification to nearly all perturbed images with an epsilon value greater than 0.1.

\begin{figure}
	\centering
	\includegraphics[scale=0.48]{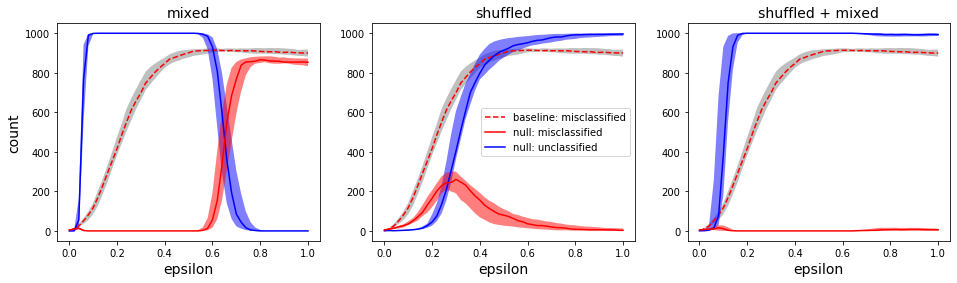}
	\caption{Comparing baseline models to best performing null models. Scores for baseline models are presented in all panels. The type of null model in each panel is indicated by the panel title. \emph{Red}: misclassified. \emph{Blue}: unclassified (a null classification). Displayed curves are the median counts across the 30 sets of perturbed images, with shaded areas indicating the 25\textsuperscript{th} and 75\textsuperscript{th} percentiles. Shaded areas are colored gray for the baseline models, for improved visual contrast with the null model results.}
	\label{fig:fig_scores_versus_epsilon_model_compare}
\end{figure}

\section{Discussion}
\label{sec:discussion}

Our findings suggest that training with a null class can provide a defense against adversarial attacks by allowing the model to give perturbed images a null classification rather than misclassifying objects within those images. For some applications this may be a perfectly acceptable behavior. A key component of this success is the use of mixed-digit null samples during training. Owing to the fact that the majority of the MNIST images have pixel values that are nearly binary, these images lie near corners of the hypercube input space. The signed gradient moves samples along the faces of the hypercube, away from the corner of the source image and toward a different corner. The mixed-digit nulls, for the most part, lie on the faces of the hypercube, between the corners, thereby forcing the model to restrict digit class decision spaces to hypercube corners. The mixed-digit null samples are particularly effective in preventing misclassification of traditional adversarial images---those with lower epsilon values and with appearances that are semantically similar or identical to those of their respective source images.

In contrast, the shuffled-digit null images are effective in preventing misclassification of \emph{high-epsilon} images as a digit class. Images perturbed in this manner are often far away from their source images in the image space, and not visually identifiable as any of the digit classes. Conventional models may nonetheless classify these images as one of the digit classes, with strong confidence. Our null-trained models rarely make this error. Like the unperturbed MNIST images, the shuffled-digit null images lie near corners of the hypercube input space. Thus their use as null samples during training can be seen as further restricting the digit class decision spaces to a smaller subset of hypercube corners.

Despite the success of these models in detecting perturbed images and classifying them as nulls, the models fail to correctly classify digits in perturbed images that are easily identifiable by humans---e.g., the majority of the blue-bordered images in Figure~\ref{fig:fig_adversarial_samples_usm}. Recent work has demonstrated that simply training a conventional model with images corrupted by Gaussian and speckle noise provides a robustness against not only those classes of noise, but to unseen types of corruption as well \cite{rusak2020increasing}. We speculate that combing such training along with the null-class training examined in this report may lead to models that increasingly perform like humans---accurately classifying perturbed images that maintain their original semantic meaning, while detecting images that have little or no semantic relation to the set of object classes and thus classifying those images as nulls.

While encouraging, the findings of this report do not ensure similar success under other models, tasks, and conditions. Further research is needed to assess the efficacy of the approach on photorealistic imagery (e.g., ImageNet) and when alternative methods of creating adversarial images are used, such as projected gradient descent \cite{madry2017towards}.

\section{Conclusion}

The conventional way in which classification models are architected and trained force the decision spaces of the classes to fill the entire input space. Our results suggest that this results in decision boundaries which can easily be crossed by adversarial examples, while maintaining an appearance and semantic meaning close to that of the unperturbed image. Including null samples during training, and removing the restriction that the model classify all possible inputs as one of the object classes, results in tighter object decision spaces, with a null space between those decision spaces.

Our approach is computationally efficient both during training and during inference, allowing for its potential use even with larger models and more challenging tasks, although its effectiveness under such conditions remains to be tested.

\bibliographystyle{unsrt}
\bibliography{references}  

\end{document}